\documentclass[conference]{IEEEtran}
\IEEEoverridecommandlockouts
\usepackage{cite}
\usepackage{amsmath,amssymb,amsfonts}
\usepackage{bm}
\usepackage{mathtools}
\usepackage{algpseudocode}
\usepackage{graphicx}
\usepackage{textcomp}
\usepackage{xcolor}
\usepackage{algorithm}
\usepackage{booktabs}
\usepackage{multirow}
\usepackage{setspace}
\usepackage{float}
\usepackage{tikz}
\usetikzlibrary{shapes,arrows.meta,positioning,patterns}
\usepackage[hidelinks]{hyperref}
\usepackage{subcaption}

\def\BibTeX{{\rm B\kern-.05em{\sc i\kern-.025em b}\kern-.08em
    T\kern-.1667em\lower.7ex\hbox{E}\kern-.125emX}}

\graphicspath{{figures/}{images/}{tables/}}

\newcommand{\x}{\mathbf{x}}

\newcommand{\W}[1]{\mathbf{W}^{#1}}

\newcommand{\R}{\mathbb{R}}

\newcommand{\relu}{\textsc{Relu}}

\renewcommand{\tanh}{\textsc{Tanh}}
\newcommand{\IfReturn}[2]{\State \algorithmicif\ #1 \algorithmicthen\ \algorithmicreturn\ #2}

\begin{document}

\title{Robustness Verification of Recurrent Neural Networks with Abstraction Refinement}

\author{\IEEEauthorblockN{Li-Jen Lin and Chih-Duo Hong\textsuperscript{*}\thanks{\textsuperscript{*}Chih-Duo Hong is the corresponding author. This research is supported by the National Science and Technology Council (NSTC), Taiwan, under grants 112-2222-E004-001-MY3 and 114-2634-F-004-002-MBK.}}
\IEEEauthorblockA{\textit{Dept. of Management Information Systems} \\
\textit{National Chengchi University}\\
Taipei, Taiwan \\
112356050@g.nccu.edu.tw, chihduo.hong@gmail.com}
}

\maketitle

\begin{abstract}
Certified local robustness verification for recurrent neural networks (RNNs) is challenging because approximation errors introduced by nonlinear relaxations can propagate through recurrent connections and accumulate over time. As a result, scalable linear bound propagation methods often become overly conservative and fail to certify inputs that are in fact robust, especially when many pre-activation intervals cross zero. We propose an abstraction-refinement framework for RNN verification that partitions such intervals to remove the dominant relaxation error: on each refined branch, \relu\ becomes exact and smooth activations such as \tanh\ and \textsc{Sigmoid} admit substantially tighter linear envelopes. To control the combinatorial cost of splitting in long sequences, we introduce a SHAP-guided timestep selection strategy that ranks hidden states by their contribution to the verification objective and refines only the most critical timesteps in temporal order. Experiments on CIFAR10 and MNIST stroke benchmarks demonstrate consistent improvements in verification success and robustness-margin tightness over abstraction-only baselines, while exposing clear runtime trade-offs between \relu\ and \tanh\ models. Data and code for reproducing our experiments can be found at \url{https://github.com/leolin-hub/ZeroSplitVerifier}.
\end{abstract}


\section{Introduction}
\label{c:intro}

Recurrent Neural Networks (RNNs) are a standard building block for learning from sequential data, underpinning applications in natural language processing~\cite{sutskever2014sequence}, speech recognition~\cite{graves2013speech}, financial forecasting~\cite{nelson2017stock}, and trajectory prediction in autonomous systems~\cite{alahi2016social}. Like other deep networks, however, RNNs can be sensitive to carefully crafted, small-magnitude input perturbations that induce incorrect predictions~\cite{szegedy2013intriguing, goodfellow2014explaining}. In critical pipelines such as perception and prediction modules for autonomous driving, these vulnerabilities can translate into high-impact failures~\cite{jia2021fooling}.

\emph{Certified robustness verification} addresses this issue by providing a formal guarantee: for a given clean sequence $\mathbf{X}_0$ with label $j$, the top-1 prediction remains unchanged for all perturbed sequences $\mathbf{X}$ within a user-specified $\ell_p$ neighborhood of radius $\varepsilon$ at each timestep.

\begin{figure}[t]
\begin{subfigure}[b]{0.45\columnwidth}
    \centering
    \begin{tikzpicture}[scale=0.6]
    \draw[->] (-3,0) -- (3,0) node[right] {$x$};
    \draw[->] (0,-1.5) -- (0,3) node[above] {$y$};
    
    \draw[thick, blue] (-2.5,0) -- (0,0) -- (2.5,2.5);
    
    \draw[dashed] (-2,0) -- (-2,2.5) node[above] {$l$};
    \draw[dashed] (2,0) -- (2,2.5) node[above] {$u$};
    
    \draw[thick, orange] (-2,0) -- (2,2);
    
    \draw[thick, red] (-2,-1) -- (2,1);
    
    \fill[orange!20, opacity=0.7] (-2,0) -- (0,0) -- (2,2) -- (-2,0);
    
    \fill[red!20, opacity=0.7] (-2,-1) -- (-2,0) -- (0,0) -- (-2,-1);
    
    \fill[red!20, opacity=0.7] (0,0) -- (2,2) -- (2,1) -- (0,0);
    
    \filldraw[black] (-2,0) circle (2pt) node[below left] {$(l,0)$};
    \filldraw[black] (2,2) circle (2pt) node[above right, yshift=-0.3cm] {$(u,u)$};
    \filldraw[black] (2,1) circle (2pt) node[below right, yshift=0.2cm] {$(u,\frac{u^2}{u-l})$};
    \filldraw[black] (-2,-1) circle (2pt) node[below left, xshift=0.4cm] {$(l,\frac{lu}{u-l})$};
    \filldraw[black] (0,0) circle (2pt);
\end{tikzpicture}
    \vspace{-1em}
    \caption{\relu}
    \label{fig:relu_approximation}
\end{subfigure}%
\hspace{-2.5em}%
\begin{subfigure}[b]{0.55\columnwidth}
    \centering
    \begin{tikzpicture}[scale=0.6]
    \draw[->] (-2.8,0) -- (2.8,0) node[right] {$x$};
    \draw[->] (0,-1.5) -- (0,3) node[above] {$y$};
    
    \draw[thick, blue, domain=-2.5:2.5, samples=100] plot (\x, {tanh(\x)});
    
    \draw[dashed] (-2,-1.2) -- (-2,2.5) node[above] {$l$};
    \draw[dashed] (2,-1.2) -- (2,2.5) node[above] {$u$};
    
    
    \pgfmathsetmacro{\tl}{tanh(-2)}  
    \pgfmathsetmacro{\tu}{tanh(2)}   
    
    \pgfmathsetmacro{\dub}{-0.76}
    \pgfmathsetmacro{\tdub}{tanh(\dub)}
    \pgfmathsetmacro{\dtdub}{1 - \tdub * \tdub}  
    \pgfmathsetmacro{\kureal}{\dtdub}
    \pgfmathsetmacro{\bureal}{\tu - \kureal * 2}
    
    \pgfmathsetmacro{\dlb}{0.76}
    \pgfmathsetmacro{\tdlb}{tanh(\dlb)}
    \pgfmathsetmacro{\dtdlb}{1 - \tdlb * \tdlb}
    \pgfmathsetmacro{\klreal}{\dtdlb}
    \pgfmathsetmacro{\blreal}{\tl - \klreal * (-2)}
    
    \draw[thick, orange, domain=-2.5:2.5] plot (\x, {\klreal * \x + \blreal});
    
    \draw[thick, red, domain=-2.5:2.5] plot (\x, {\kureal * \x + \bureal});
    
    \fill[red!20, opacity=0.5, domain=-2:2, samples=50] 
        plot (\x, {tanh(\x)}) -- 
        plot[domain=2:-2] (\x, {\kureal * \x + \bureal}) -- 
        cycle;
    
    \fill[orange!20, opacity=0.5, domain=-2:2, samples=50] 
        plot (\x, {\klreal * \x + \blreal}) -- 
        plot[domain=2:-2] (\x, {tanh(\x)}) -- 
        cycle;
    
    \filldraw[black] (-2,\tl) circle (2pt) node[above left, font=\small] {$(l,\textrm{tanh}(l))$};
    \filldraw[black] (2,\tu) circle (2pt) node[below right, font=\small] {$(u,\textrm{tanh}(u))$};
    \filldraw[black] (0,0) circle (2pt);
    
    \filldraw[red] (\dub,\tdub) circle (2pt) node[below right, font=\scriptsize, xshift=-0.2cm] {$d_{LB}$};
    \filldraw[orange] (\dlb,\tdlb) circle (2pt) node[above left, font=\scriptsize, xshift=0.2cm] {$d_{UB}$};
\end{tikzpicture}
    \vspace{-.8em}
    \caption{\tanh}
    \label{fig:tanh_approximation}
\end{subfigure}
\caption{Linear relaxation for \relu\ and \tanh\ when the pre-activation interval $[l,u]$ straddles zero. The shaded regions represent the over-approximation introduced by the relaxation.}
\label{fig:activation_approximation}
\end{figure}
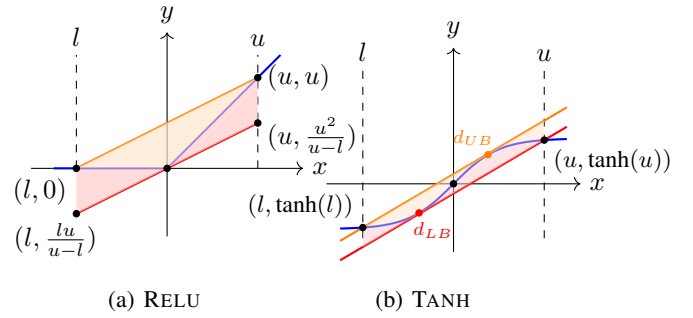

Despite recent progress, producing \emph{tight} certificates for RNNs remains difficult. The strongest scalable verifiers typically rely on abstraction-based bound propagation (e.g., linear bound propagation for RNNs~\cite{ko2019popqorn,li2023towards}), which replaces nonlinear activations with linear relaxations to obtain certified output bounds. The main failure mode arises when a neuron's \emph{pre-activation} interval crosses zero. In this regime, the relaxation gap can be large for both piecewise-linear activations (\relu) and sigmoid activations (\tanh), see Figure~\ref{fig:activation_approximation}. In an RNN, that over-approximation error is fed back through recurrent connections and can accumulate across timesteps, yielding conservative bounds and frequent \emph{false negatives} (robust inputs that cannot be certified).

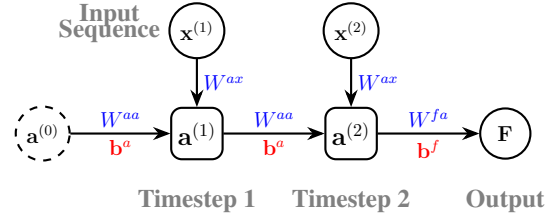
\begin{figure}[t]
\centering
\begin{tikzpicture}[
    scale=0.68,
    transform shape,
    neuron/.style={circle, draw, thick, minimum size=0.6cm, font=\large},
    hidden/.style={rectangle, draw, thick, rounded corners, minimum size=1cm, font=\Large},
    arrow/.style={-Stealth, thick},
    weight/.style={font=\large, text=blue},
    diagramlabel/.style={font=\Large\bfseries}
]

\node[neuron, dashed] (h0) at (0,0) {$\mathbf{a}^{(0)}$};

\node[hidden] (h1) at (3,0) {$\mathbf{a}^{(1)}$};
\node[neuron] (x1) at (3,2) {$\mathbf{x}^{(1)}$};
\node[diagramlabel, text=gray] at (3,-1.3) {Timestep 1};  

\node[hidden] (h2) at (6,0) {$\mathbf{a}^{(2)}$};
\node[neuron] (x2) at (6,2) {$\mathbf{x}^{(2)}$};
\node[diagramlabel, text=gray] at (6,-1.3) {Timestep 2};  

\node[neuron] (output) at (9,0) {$~\mathbf{F}~$};
\node[diagramlabel, text=gray] at (9,-1.3) {Output};  

\draw[arrow] (h0) -- node[above, weight] {$W^{aa}$} node[below, weight, text=red] {$\mathbf{b}^a$} (h1);
\draw[arrow] (h1) -- node[above, weight] {$W^{aa}$} node[below, weight, text=red] {$\mathbf{b}^a$} (h2);

\draw[arrow] (x1) -- node[right, weight] {$W^{ax}$} (h1);
\draw[arrow] (x2) -- node[right, weight] {$W^{ax}$} (h2);

\draw[arrow] (h2) -- node[above, weight] {$W^{fa}$} node[below, weight, text=red] {$\mathbf{b}^f$} (output);

\node[font=\footnotesize, text=purple] at (3.3,-0.65) {};
\node[font=\footnotesize, text=purple] at (6.3,-0.65) {};

\node[diagramlabel, text=gray] at (1.3,2.3) {Input};
\node[diagramlabel, text=gray] at (1.3,2.0) {Sequence};

\end{tikzpicture}
\caption{A vanilla RNN architecture with 2 timesteps.} 
\label{fig:rnn_simple}
\end{figure}

Abstraction refinement is a natural way to reduce this conservatism. When the initial abstraction is too coarse, the verifier can branch on unstable pre-activation regions and solve smaller sub-problems with tighter activation bounds. For RNNs, however, refinement is not just a spatial branching problem: splitting one hidden neuron at timestep $t$ changes the feasible hidden-state set at that timestep and therefore affects every later step in the unrolled recurrence. Moreover, naively splitting all cross-zero neurons quickly becomes intractable because the number of sub-problems grows exponentially.

This paper develops a \emph{single-neuron} refinement workflow tailored to RNN verification. When abstraction-only bound propagation fails, we identify a small set of critical cross-zero neurons and perform refinement one neuron at a time. For a selected neuron $(t,r)$, we create a positive branch by constraining its pre-activation interval to $[0, u_r^{(t)}]$ and a negative branch by constraining it to $[l_r^{(t)}, 0]$, then re-propagate bounds through the remaining timesteps and verify each resulting sub-task. For \relu, this removes the dominant cross-zero relaxation error exactly on the split neuron; for \tanh, it yields a substantially tighter local envelope on each branch.

To decide where to refine, we use gradient-based SHAP attributions on hidden states to rank individual neuron instances by their contribution to the verification objective. We restrict attention to neurons whose pre-activation bounds still cross zero, sort the highest-scoring candidates in temporal order, and refine them under a depth budget. This yields a targeted search strategy that preserves the temporal leverage of early splits without requiring exhaustive branching over all neurons or all timesteps.

We evaluate our approach on two temporal benchmarks (CIFAR10 reshaped as sequences and MNIST stroke trajectories) using vanilla RNN classifiers with \relu\ and \tanh\ activations. Across model sizes, sequence lengths, and perturbation radii, SHAP-guided single-neuron refinement increases the number of verified samples and consistently tightens certified robustness margins compared to abstraction-only baselines, with predictable compute overhead that grows with sequence length and split depth.

\textit{Contributions.} Our main contributions in this paper are:
(1) We develop a splitting-based abstraction refinement workflow for RNN verification that propagates each split through the unrolled recurrence.
(2) We introduce a SHAP-guided neuron ranking strategy that prioritizes high-impact candidate splits while respecting temporal dependencies.
(3) We provide an empirical evaluation on two sequence benchmarks demonstrating improved certification rates and tighter output bounds, together with an analysis of runtime trade-offs across activations and sequence lengths.

\textit{Organization.}
The rest of this paper is organized as follows.
Section~\ref{c:abstract_refinement} reviews robustness verification for RNNs via abstraction-based bound propagation.
Section~\ref{c:zerosplit} presents our abstraction-refinement framework, including the splitting rule and SHAP-guided neuron selection strategy.
Section~\ref{c:experiment} reports the experimental evaluation and discusses threats to validity.
Section~\ref{c:related} discusses related work and closes with concluding remarks.

\section{Robustness Verification via Abstraction}
\label{c:abstract_refinement}
   Consider an RNN classifier $F: \R^{n \times m} \rightarrow \R^c$, where $n$ is the per-timestep input dimension, $m$ is the sequence length, and $c$ is the number of classes. For an input sequence $\mathbf{X}_0 = [\x_0^{(1)}, \ldots, \x_0^{(m)}]$ with ground-truth label $j$, our goal is to verify that the top-1 classification remains unchanged under input perturbations. The adversarially perturbed sequence is denoted as $\mathbf{X} = [\x^{(1)}, \ldots, \x^{(m)}]$, where each perturbed frame $\x^{(k)}$ is constrained within an $\varepsilon$-radius $l_p$-ball: $\x^{(k)} \in \mathbb{B}_p(\x_0^{(k)}, \varepsilon)$. Specifically, we want to certify that the ground-truth logit remains dominant for all such perturbed inputs; a sufficient condition is $\gamma_j^L > \gamma_i^U$ for all $i \neq j$.

  We tackle this verification problem by computing certified lower and upper bounds $\gamma_j^L$ and $\gamma_i^U$ such that $\gamma_j^L > \gamma_i^U$ for all classes $i \neq j$. The challenge lies in computing tight enough bounds to make this verification feasible, as loose bounds often lead to false negatives where actually robust models fail verification.
  
  We conduct the verification through a two-phase procedure. First, we use bound propagation to establish initial certified robustness bounds across RNN layers, leveraging output bounds estimation from \textsc{Popqorn}~\cite{ko2019popqorn}. We refer to the forward computation of hidden pre-activation bounds as \emph{abstract bound propagation (ABP)} and to the subsequent computation of certified output bounds as \emph{backward bound propagation (BBP)}. If these initial bounds fail to verify the desired robustness, we proceed to apply abstraction refinement (Section~\ref{c:zerosplit}) to improve the precision of the bounds.
  
  The verification procedure first derives input-dependent linear bounds $F_j^{L}(\mathbf{X})$ and $F_j^{U}(\mathbf{X})$ for each logit $F_j$, then optimizes them over the perturbation set to obtain scalar certificates $\gamma_j^L$ and $\gamma_j^U$. Concretely, for any perturbed input sequence $\mathbf{X} = [\x^{(1)}, \ldots, \x^{(m)}]$ with $\x^{(k)}\in\mathbb{B}_p(\x_0^{(k)},\varepsilon)$ for $k \in \{1,\dots,m\}$, BBP constructs $F_j^{L}(\mathbf{X})$ and $F_j^{U}(\mathbf{X})$ so that
\begin{align*}
F_j^{L}(\mathbf{X})\leq F_j(\mathbf{X})\leq F_j^{U}(\mathbf{X}).
\end{align*}
In the following, we show the derivation for the 2-timestep many-to-one RNN in Figure~\ref{fig:rnn_simple}. The general $m$-timestep case follows by the same backward recursion.
\begin{align*}
F(\mathbf{X}) &= \W{fa} \mathbf{a}^{(2)}+\mathbf{b}^{f}, \\
\mathbf{a}^{(2)} &= \sigma(\mathbf{y}^{(2)}),\quad \mathbf{y}^{(2)} = \W{aa} \mathbf{a}^{(1)}+\W{ax} \x^{(2)}+\mathbf{b}^{a}, \\
\mathbf{a}^{(1)} &= \sigma(\mathbf{y}^{(1)}),\quad \mathbf{y}^{(1)} = \W{aa} \mathbf{a}^{(0)}+\W{ax} \x^{(1)}+\mathbf{b}^{a}. 
\end{align*}
Here $\mathbf{a}^{(0)}\in\mathbb{R}^s$ is the initial hidden state, $\sigma$ is the hidden activation function, and $\mathbf{y}^{(t)}\in\mathbb{R}^s$ is the pre-activation vector at timestep $t$. ABP computes sound pre-activation bounds $\mathbf{l}^{(t)},\mathbf{u}^{(t)}\in\mathbb{R}^s$ with coordinates $l_r^{(t)}\le y_r^{(t)}\le u_r^{(t)}$. We index a hidden neuron by $(t,r)$, meaning neuron $r$ at timestep $t$, and call it \emph{cross-zero} when $l_r^{(t)}<0<u_r^{(t)}$.

For each interval $[l_r^{(t)},u_r^{(t)}]$, BBP uses affine lower and upper bounds on $\sigma$:
\begin{align}
h_{L,r}^{(t)}(v)=\alpha_{L,r}^{(t)}v+\beta_{L,r}^{(t)},\quad
h_{U,r}^{(t)}(v)=\alpha_{U,r}^{(t)}v+\beta_{U,r}^{(t)},
\label{eq:act_linear_bounds}
\end{align}
such that $h_{L,r}^{(t)}(v)\le \sigma(v)\le h_{U,r}^{(t)}(v)$ for all $v\in[l_r^{(t)},u_r^{(t)}]$. For {\relu}, the non-cross-zero cases are exact: if $l_r^{(t)}\ge 0$, then $(\alpha_{L,r}^{(t)},\beta_{L,r}^{(t)})=(\alpha_{U,r}^{(t)},\beta_{U,r}^{(t)})=(1,0)$; if $u_r^{(t)}\le 0$, then both pairs equal $(0,0)$. If $l_r^{(t)}<0<u_r^{(t)}$, then
\begin{align*}
\alpha_{U,r}^{(t)}=\alpha_{L,r}^{(t)}=\frac{u_r^{(t)}}{u_r^{(t)}-l_r^{(t)}},\quad
\beta_{L,r}^{(t)}=0,\quad
\beta_{U,r}^{(t)}=-\frac{l_r^{(t)}u_r^{(t)}}{u_r^{(t)}-l_r^{(t)}}.
\end{align*}
For {\tanh} and \textsc{Sigmoid}, we analogously use valid affine envelopes over $[l_r^{(t)},u_r^{(t)}]$.

To upper-bound the logit of class $j$, start from the output layer:
\begin{align*}
F_j(\mathbf{X}) = \W{fa}_{j,:}\sigma(\mathbf{y}^{(2)})+b_j^{f}.
\end{align*}
Let $\mathbf{c}_{j,:}^{(2)}\coloneqq \W{fa}_{j,:}$ denote the current backward coefficient row on $\mathbf{a}^{(2)}$, and let $c_{j,r}^{(t)}$ denote the $r$-th entry of $\mathbf{c}_{j,:}^{(t)}$. To preserve an upper bound, each coordinate chooses the upper or lower activation line according to the sign of $c_{j,r}^{(t)}$:
\begin{align*}
\lambda_{j,r}^{(t)}&=
\begin{cases}
\alpha_{U,r}^{(t)}, & c_{j,r}^{(t)}\ge 0,\\
\alpha_{L,r}^{(t)}, & c_{j,r}^{(t)}<0;
\end{cases}
&
\delta_{r,j}^{(t)}&=
\begin{cases}
\beta_{U,r}^{(t)}, & c_{j,r}^{(t)}\ge 0,\\
\beta_{L,r}^{(t)}, & c_{j,r}^{(t)}<0.
\end{cases}
\end{align*}
Collect $\lambda_{j,r}^{(t)}$ into the row vector $\lambda_{j,:}^{(t)}\in\mathbb{R}^s$ and $\delta_{r,j}^{(t)}$ into the column vector $\delta_{:,j}^{(t)}\in\mathbb{R}^s$. Define $\Lambda_{j,:}^{(t)}\coloneqq \mathbf{c}_{j,:}^{(t)}\odot\lambda_{j,:}^{(t)}$, where $\odot$ denotes elementwise product, and let $\eta_j^{(t)}\coloneqq \langle\mathbf{c}_{j,:}^{(t)},\delta_{:,j}^{(t)}\rangle$ be the scalar intercept contribution. Applying Eq.~\eqref{eq:act_linear_bounds} at timestep $2$ gives
\begin{align*}
F_j(\mathbf{X})
&\le \Lambda_{j,:}^{(2)}\mathbf{y}^{(2)}+\eta_j^{(2)}+b_j^{f} = \Lambda_{j,:}^{(2)}\W{aa}\mathbf{a}^{(1)} + \nonumber\\
&\qquad \Lambda_{j,:}^{(2)}\W{ax}\x^{(2)} + \Lambda_{j,:}^{(2)}\mathbf{b}^{a} + \eta_j^{(2)} + b_j^{f}.
\end{align*}
Now set $\mathbf{c}_{j,:}^{(1)}\coloneqq \Lambda_{j,:}^{(2)}\W{aa}$ and apply the same construction to $\mathbf{a}^{(1)}=\sigma(\mathbf{y}^{(1)})$. After this second backward step, we obtain the linear upper bound
\begin{align*}
F_j(\mathbf{X})
&\;\le\; \Lambda_{j,:}^{(1)}\W{aa}\mathbf{a}^{(0)} + \sum_{z=1}^{2}\Lambda_{j,:}^{(z)}\W{ax}\x^{(z)} \nonumber\\
&\qquad + \sum_{z=1}^{2}\big(\Lambda_{j,:}^{(z)}\mathbf{b}^{a}+\eta_j^{(z)}\big) + b_j^{f}.
\end{align*}
We then set the RHS of the above equation to $F_j^{U}(\mathbf{X})$. Since $F_j^{U}(\mathbf{X})$ is linear in each perturbed frame $\x^{(z)}$, maximizing it over $\x^{(z)}\in\mathbb{B}_p(\x_0^{(z)},\varepsilon)$ and using H\"older's inequality yields the certified scalar upper bound
\begin{align*}
\gamma_j^{U}
&= \Lambda_{j,:}^{(1)}\W{aa}\mathbf{a}^{(0)} + \sum_{z=1}^{2}\Big(\Lambda_{j,:}^{(z)}\W{ax}\x_0^{(z)} + \varepsilon\|\Lambda_{j,:}^{(z)}\W{ax}\|_{q} \nonumber\\
&\qquad\qquad\qquad\; + \Lambda_{j,:}^{(z)}\mathbf{b}^{a}+\eta_j^{(z)}\Big) + b_j^{f},
\end{align*}
where $q$ is the dual norm of $p$, i.e., $1/p+1/q=1$.

We now derive the certified lower bound in the same 2-timestep setting. BBP maintains a separate lower-bound backward coefficient row $\widetilde{\mathbf{c}}_{j,:}^{(t)}$, initialized at the output layer by $\widetilde{\mathbf{c}}_{j,:}^{(2)}\coloneqq \W{fa}_{j,:}$, and we write $\widetilde{c}_{j,r}^{(t)}$ for its $r$-th entry. To preserve a lower bound, each coordinate chooses the lower or upper activation line according to the sign of $\widetilde{c}_{j,r}^{(t)}$:
\begin{align*}
\lambda_{j,r}^{L,(t)}&=
\begin{cases}
\alpha_{L,r}^{(t)}, & \widetilde{c}_{j,r}^{(t)}\ge 0,\\
\alpha_{U,r}^{(t)}, & \widetilde{c}_{j,r}^{(t)}<0;
\end{cases}
&
\delta_{r,j}^{L,(t)}&=
\begin{cases}
\beta_{L,r}^{(t)}, & \widetilde{c}_{j,r}^{(t)}\ge 0,\\
\beta_{U,r}^{(t)}, & \widetilde{c}_{j,r}^{(t)}<0.
\end{cases}
\end{align*}
Collect $\lambda_{j,r}^{L,(t)}$ into the row vector $\lambda_{j,:}^{L,(t)}\in\mathbb{R}^s$ and $\delta_{r,j}^{L,(t)}$ into the column vector $\delta_{:,j}^{L,(t)}\in\mathbb{R}^s$. Define $\Lambda_{j,:}^{L,(t)}\coloneqq \widetilde{\mathbf{c}}_{j,:}^{(t)}\odot\lambda_{j,:}^{L,(t)}$ and $\eta_j^{L,(t)}\coloneqq \langle \widetilde{\mathbf{c}}_{j,:}^{(t)},\delta_{:,j}^{L,(t)}\rangle$. Applying Eq.~\eqref{eq:act_linear_bounds} at timestep $2$ gives
\begin{align*}
F_j(\mathbf{X})
&\ge \Lambda_{j,:}^{L,(2)}\mathbf{y}^{(2)}+\eta_j^{L,(2)}+b_j^{f} = \Lambda_{j,:}^{L,(2)}\W{aa}\mathbf{a}^{(1)} + \nonumber\\
&\qquad \Lambda_{j,:}^{L,(2)}\W{ax}\x^{(2)} + \Lambda_{j,:}^{L,(2)}\mathbf{b}^{a} + \eta_j^{L,(2)} + b_j^{f}.
\end{align*}
Now set $\widetilde{\mathbf{c}}_{j,:}^{(1)}\coloneqq \Lambda_{j,:}^{L,(2)}\W{aa}$ and apply the same construction to $\mathbf{a}^{(1)}=\sigma(\mathbf{y}^{(1)})$. After this second backward step, we obtain the linear lower bound
\begin{align*}
F_j(\mathbf{X})
&\;\ge\; \Lambda_{j,:}^{L,(1)}\W{aa}\mathbf{a}^{(0)} + \sum_{z=1}^{2}\Lambda_{j,:}^{L,(z)}\W{ax}\x^{(z)} \nonumber\\
&\qquad + \sum_{z=1}^{2}\big(\Lambda_{j,:}^{L,(z)}\mathbf{b}^{a}+\eta_j^{L,(z)}\big) + b_j^{f}.
\end{align*}
We then set the RHS of the above equation to $F_j^{L}(\mathbf{X})$. Since $F_j^{L}(\mathbf{X})$ is linear in each perturbed frame $\x^{(z)}$, minimizing it over $\x^{(z)}\in\mathbb{B}_p(\x_0^{(z)},\varepsilon)$ and using H\"older's inequality yields the certified scalar lower bound
\begin{align*}
\gamma_j^{L}
&= \Lambda_{j,:}^{L,(1)}\W{aa}\mathbf{a}^{(0)} + \sum_{z=1}^{2}\Big(\Lambda_{j,:}^{L,(z)}\W{ax}\x_0^{(z)} - \varepsilon\|\Lambda_{j,:}^{L,(z)}\W{ax}\|_{q} \nonumber\\
&\qquad\qquad\qquad\qquad\quad\qquad\; + \Lambda_{j,:}^{L,(z)}\mathbf{b}^{a}+\eta_j^{L,(z)}\Big) + b_j^{f}.
\end{align*}

\smallskip
\textbf{Example.} We use the 2-timestep RNN in Figure~\ref{fig:rnn_simple} as a running example. Consider a sample with ground-truth label $j$ and perturbation radius $\varepsilon=0.05$. Suppose ABP yields two cross-zero neurons: neuron $(1,2)$, i.e., neuron $2$ at timestep $1$, with $l_2^{(1)}=-0.04$ and $u_2^{(1)}=0.09$, and neuron $(2,4)$ with $l_4^{(2)}=-0.02$ and $u_4^{(2)}=0.05$. Neuron $(1,2)$ is more harmful because any relaxation error introduced at timestep~1 is propagated through the recurrent transition before the output is computed. After running ABP followed by BBP, assume we obtain $\gamma^L_j=0.18$ and $\max_{i\neq j}\gamma^U_i=0.21$, so abstraction alone cannot certify the sample. Section~\ref{c:zerosplit} revisits this same sample and shows how splitting only one neuron can recover the certificate.

\section{Abstraction Refinement}
\label{c:zerosplit}

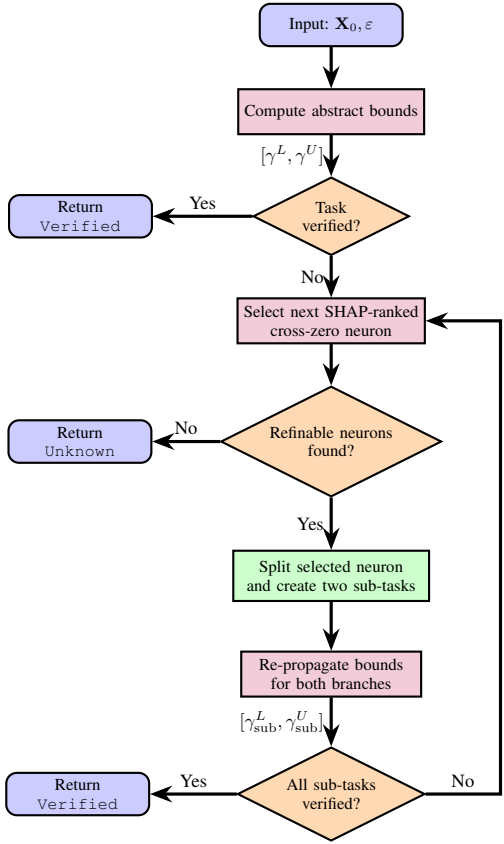
\begin{figure}[!t]
\centering
\begin{tikzpicture}[scale=0.7, every node/.style={scale=0.7},
    node distance=0.95cm,
    startstop/.style={rectangle, rounded corners, minimum width=2.7cm, minimum height=0.8cm,
                      text centered, draw=black, fill=blue!20, thick, font=\small, align=center},
    process/.style={rectangle, minimum width=3.4cm, minimum height=0.85cm,
                    text centered, draw=black, fill=purple!20, thick, font=\small, align=center},
    zeroprocess/.style={rectangle, minimum width=3.7cm, minimum height=0.95cm,
                        text centered, draw=black, fill=green!20, thick, font=\small, align=center},
    decision/.style={diamond, aspect=2.0, minimum width=2.3cm, minimum height=0.9cm,
                     text centered, draw=black, fill=orange!30, thick, font=\small, align=center},
    arrow/.style={very thick, -Stealth}
]

\node (start) [startstop] at (0,0) {Input: $\mathbf{X}_0,\varepsilon$};
\node (bounds1) [process, below=.55cm of start] {Compute abstract bounds};
\node (check1) [decision, below=.55cm of bounds1] {Task\\verified?};
\node (safe1) [startstop, left=1.32cm of check1] {Return\\\texttt{Verified}};

\node (select) [process, below=.55cm of check1] {Select next SHAP-ranked\\cross-zero neuron};
\node (check2) [decision, below=.55cm of select] {Refinable neurons\\found?}; 
\node (unknown) [startstop, left=0.9cm of check2] {Return\\\texttt{Unknown}};
\node (split) [zeroprocess, below=.45cm of check2, yshift=-0.35cm]
    {Split selected neuron\\and create two sub-tasks};
\node (bounds2) [process, below=.65cm of split] {Re-propagate bounds\\for both branches};
\node (check3) [decision, below=.65cm of bounds2] {All sub-tasks\\verified?};
\node (safe2) [startstop, left=1.15cm of check3] {Return\\\texttt{Verified}};

\draw [arrow] (start) -- (bounds1);
\draw [arrow] (bounds1) -- node[anchor=east] {$[\gamma^L,\gamma^U]$} (check1);
\draw [arrow] (check1) -- node[anchor=south] {Yes} (safe1);
\draw [arrow] (check1) -- node[anchor=east] {No} (select);

\draw [arrow] (select) -- (check2);
\draw [arrow] (check2) -- node[anchor=south] {No} (unknown);
\draw [arrow] (check2) -- node[anchor=east] {Yes} (split);
\draw [arrow] (split) -- (bounds2);
\draw [arrow] (bounds2) -- node[anchor=east] {$[\gamma^L_{\mathrm{sub}},\gamma^U_{\mathrm{sub}}]$} (check3);
\draw [arrow] (check3) -- node[anchor=south] {Yes} (safe2);
\draw [arrow] (check3) -| node[near start, anchor=south] {No} ++(3.2,0) |- (select);

\end{tikzpicture}
\caption{Verification workflow for local robustness of RNN}
\label{fig:zerosplit_framework}
\end{figure}

Linear bound propagation can become overly conservative when many pre-activations cross zero, where activation relaxations are loosest. In RNNs, these local errors are amplified through the recurrent map across timesteps, yielding false negatives (uncertified yet actually robust samples). We mitigate this by splitting pre-activation ranges that cross zero, which produce sub-problems with exact {\relu} behavior and substantially tighter {\tanh} or \textsc{Sigmoid} envelopes.

\paragraph{Verification workflow}
As shown in Figure~\ref{fig:zerosplit_framework} and Algorithm~\ref{alg:overall_refinement}, verification starts from the perturbed input instance $(\mathbf{X}_0,\varepsilon)$ by computing certified abstract output bounds $[\gamma^L,\gamma^U]$ via abstract bound propagation followed by backward bound propagation. If the robustness margin holds (i.e., the lower bound of the ground-truth class dominates all competing upper bounds), the procedure terminates and returns \texttt{Verified}. Otherwise, we search for a \emph{critical neuron} whose pre-activation bound still crosses zero (selected from the SHAP-ranked candidate set). If no such neuron remains (or refinement cannot proceed under the depth budget), the result is returned as \texttt{Unknown}. When a critical neuron $r^*$ at timestep $t_c$ is found, we perform a zero-split on $r^*$ to create positive/negative sub-tasks by constraining that neuron to $[0,u_{r^*}^{(t_c)}]$ or $[l_{r^*}^{(t_c)},0]$, respectively, then recompute abstract bounds for each sub-task by propagating the refined constraint through the remaining unrolled timesteps. The overall task is certified only if \emph{all} generated sub-tasks satisfy the margin; otherwise, the loop repeats by selecting the next critical neuron and further refining the remaining failing branches.

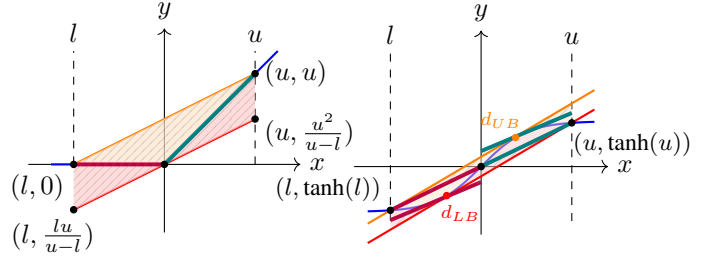
\begin{figure}[t]
\vspace{-0.5em}
\centering
\hspace{-7em}
\begin{subfigure}[b]{0.40\columnwidth}
    \centering
    \begin{tikzpicture}[scale=0.6]
    \draw[->] (-3,0) -- (3,0) node[right] {$x$};
    \draw[->] (0,-1.5) -- (0,3) node[above] {$y$};
    
    \draw[thick, blue] (-2.5,0) -- (0,0) -- (2.5,2.5);
    
    \draw[dashed] (-2,0) -- (-2,2.5) node[above] {$l$};
    \draw[dashed] (2,0) -- (2,2.5) node[above] {$u$};
    
    \draw[thick, orange] (-2,0) -- (2,2);
    
    \draw[thick, red] (-2,-1) -- (2,1);
    
    \fill[orange!20, opacity=0.7] (-2,0) -- (0,0) -- (2,2) -- (-2,0);
    \fill[red!20, opacity=0.7] (-2,-1) -- (-2,0) -- (0,0) -- (-2,-1);
    \fill[red!20, opacity=0.7] (0,0) -- (2,2) -- (2,1) -- (0,0);
    
    \draw[line width=1.5pt, purple] (-2,0) -- (0,0); 
    \draw[line width=1.5pt, teal] (0,0) -- (2,2);  

    
    \fill[pattern=north east lines, pattern color=gray, opacity=0.5] 
        (-2,0) -- (0,1) -- (0,0) -- cycle;
        
    \fill[pattern=north east lines, pattern color=gray, opacity=0.5] 
        (-2,-1) -- (0,0) -- (-2,0) -- cycle;

    \fill[pattern=north east lines, pattern color=gray, opacity=0.5] 
        (0,1) -- (2,2) -- (0,0) -- cycle;
        
    \fill[pattern=north east lines, pattern color=gray, opacity=0.5] 
        (0,0) -- (2,1) -- (2,2) -- cycle;

    \filldraw[black] (-2,0) circle (2pt) node[below left] {$(l,0)$};
    \filldraw[black] (2,2) circle (2pt) node[above right, yshift=-0.3cm] {$(u,u)$};
    \filldraw[black] (2,1) circle (2pt) node[below right, yshift=0.2cm] {$(u,\frac{u^2}{u-l})$};
    \filldraw[black] (-2,-1) circle (2pt) node[below left, xshift=0.4cm] {$(l,\frac{lu}{u-l})$};
    \filldraw[black] (0,0) circle (2pt);
\end{tikzpicture}
\end{subfigure}
\hspace{-.8em}
\begin{subfigure}[b]{0.40\columnwidth}
    \centering
    \begin{tikzpicture}[scale=0.6]
    \draw[->] (-2.8,0) -- (2.8,0) node[right] {$x$};
    \draw[->] (0,-1.5) -- (0,3) node[above] {$y$};
    
    \draw[thick, blue, domain=-2.5:2.5, samples=100] plot (\x, {tanh(\x)});
    
    \draw[dashed] (-2,-1.2) -- (-2,2.5) node[above] {$l$};
    \draw[dashed] (2,-1.2) -- (2,2.5) node[above] {$u$};
    
    \pgfmathsetmacro{\tl}{tanh(-2)}
    \pgfmathsetmacro{\tu}{tanh(2)}
    
    \pgfmathsetmacro{\dub}{-0.76}
    \pgfmathsetmacro{\tdub}{tanh(\dub)}
    \pgfmathsetmacro{\dtdub}{1 - \tdub * \tdub}
    \pgfmathsetmacro{\kureal}{\dtdub}
    \pgfmathsetmacro{\bureal}{\tu - \kureal * 2}
    
    \pgfmathsetmacro{\dlb}{0.76}
    \pgfmathsetmacro{\tdlb}{tanh(\dlb)}
    \pgfmathsetmacro{\dtdlb}{1 - \tdlb * \tdlb}
    \pgfmathsetmacro{\klreal}{\dtdlb}
    \pgfmathsetmacro{\blreal}{\tl - \klreal * (-2)}
    
    \pgfmathsetmacro{\tanhOne}{tanh(1)}
    \pgfmathsetmacro{\slopePos}{1 - \tanhOne*\tanhOne}
    \pgfmathsetmacro{\interceptPos}{\tanhOne - \slopePos}
    
    \pgfmathsetmacro{\slopeNeg}{\slopePos}
    \pgfmathsetmacro{\interceptNeg}{-\interceptPos}

    \fill[red!20, opacity=0.5, domain=-2:2, samples=50] 
        plot (\x, {tanh(\x)}) -- 
        plot[domain=2:-2] (\x, {\kureal * \x + \bureal}) -- cycle;
    
    \fill[orange!20, opacity=0.5, domain=-2:2, samples=50] 
        plot (\x, {\klreal * \x + \blreal}) -- 
        plot[domain=2:-2] (\x, {tanh(\x)}) -- cycle;

    \draw[thick, orange, domain=-2.5:2.5] plot (\x, {\klreal * \x + \blreal});
    \draw[thick, red, domain=-2.5:2.5] plot (\x, {\kureal * \x + \bureal});

    \draw[line width=1.5pt, teal] (0,0) -- (2, \tu); 
    \draw[line width=1.5pt, teal] (0, \interceptPos) -- (2, {\slopePos*2 + \interceptPos}); 

    \draw[line width=1.5pt, purple] (-2, \tl) -- (0,0); 
    \draw[line width=1.5pt, purple] (-2, {\slopeNeg*(-2) + \interceptNeg}) -- (0, \interceptNeg); 

    
    \fill[pattern=north east lines, pattern color=gray, opacity=0.5]
        (0, \blreal) -- (2, {\klreal*2 + \blreal}) -- 
        (2, {\slopePos*2 + \interceptPos}) -- (0, \interceptPos) -- cycle;

    \fill[pattern=north east lines, pattern color=gray, opacity=0.5]
        (0, \bureal) -- (2, {\kureal*2 + \bureal}) -- 
        (2, \tu) -- (0, 0) -- cycle;
        
    \fill[pattern=north east lines, pattern color=gray, opacity=0.5]
        (-2, {\klreal*(-2) + \blreal}) -- (0, \blreal) --
        (0, 0) -- (-2, \tl) -- cycle;

    \fill[pattern=north east lines, pattern color=gray, opacity=0.5]
        (-2, {\kureal*(-2) + \bureal}) -- (0, \bureal) --
        (0, \interceptNeg) -- (-2, {\slopeNeg*(-2) + \interceptNeg}) -- cycle;

    \filldraw[black] (-2,\tl) circle (2pt) node[above left, font=\small] {$(l,\textrm{tanh}(l))$};
    
    \filldraw[black] (2,\tu) circle (2pt) node[below right, xshift=-3pt, font=\small] {$(u,\textrm{tanh}(u))$};
    
    \filldraw[black] (0,0) circle (2pt);
    
    \filldraw[red] (\dub,\tdub) circle (2pt) node[below right, font=\scriptsize, xshift=-0.2cm] {$d_{LB}$};
    \filldraw[orange] (\dlb,\tdlb) circle (2pt) node[above left, font=\scriptsize, xshift=0.2cm] {$d_{UB}$};
\end{tikzpicture}
    \vspace{.7em}
\end{subfigure}
\vspace{-1em}
\caption{Splitting the pre-activation interval at zero tightens the relaxation (which becomes exact for {\relu}).}
\label{fig:activation_refinement}
\vspace{-0.7em}
\end{figure}

\paragraph{Split rule}
Given a timestep $t_c$ and a selected cross-zero neuron $r^*\in\mathcal{I}_{\text{cross}}(t_c)$, where
$\mathcal{I}_{\text{cross}}(t_c)=\{r \mid l_r^{(t_c)}<0<u_r^{(t_c)}\}$,
let $(\mathbf{l}_{\text{pos}}^{(t)},\mathbf{u}_{\text{pos}}^{(t)})$ and $(\mathbf{l}_{\text{neg}}^{(t)},\mathbf{u}_{\text{neg}}^{(t)})$ denote the branch-specific pre-activation bounds after splitting. We create two branches by constraining only the selected neuron:
(i) the \emph{positive} branch sets $l_{\text{pos},r^*}^{(t_c)}\!\leftarrow\!0$ and
(ii) the \emph{negative} branch sets $u_{\text{neg},r^*}^{(t_c)}\!\leftarrow\!0$.
For {\relu}, each branch is exact; for {\tanh}, restricting the domain yields a tighter linear envelope (Figure~\ref{fig:activation_refinement}). Each branch produces its own certified output bounds $[\gamma^L,\gamma^U]$. The original sample is certified iff every branch satisfies
\begin{equation}
\label{eq:verify_margin}
\gamma^L_j > \max\nolimits_{i \neq j}\gamma^U_i\!,\quad\text{$j$ is the top-1 class label.}
\end{equation}
This is the \textsc{Split} step in Algorithm~\ref{alg:zerosplit}.

\begin{figure}[!t]
\vspace{-0.7em}
\centering
\includegraphics[width=0.82\columnwidth]{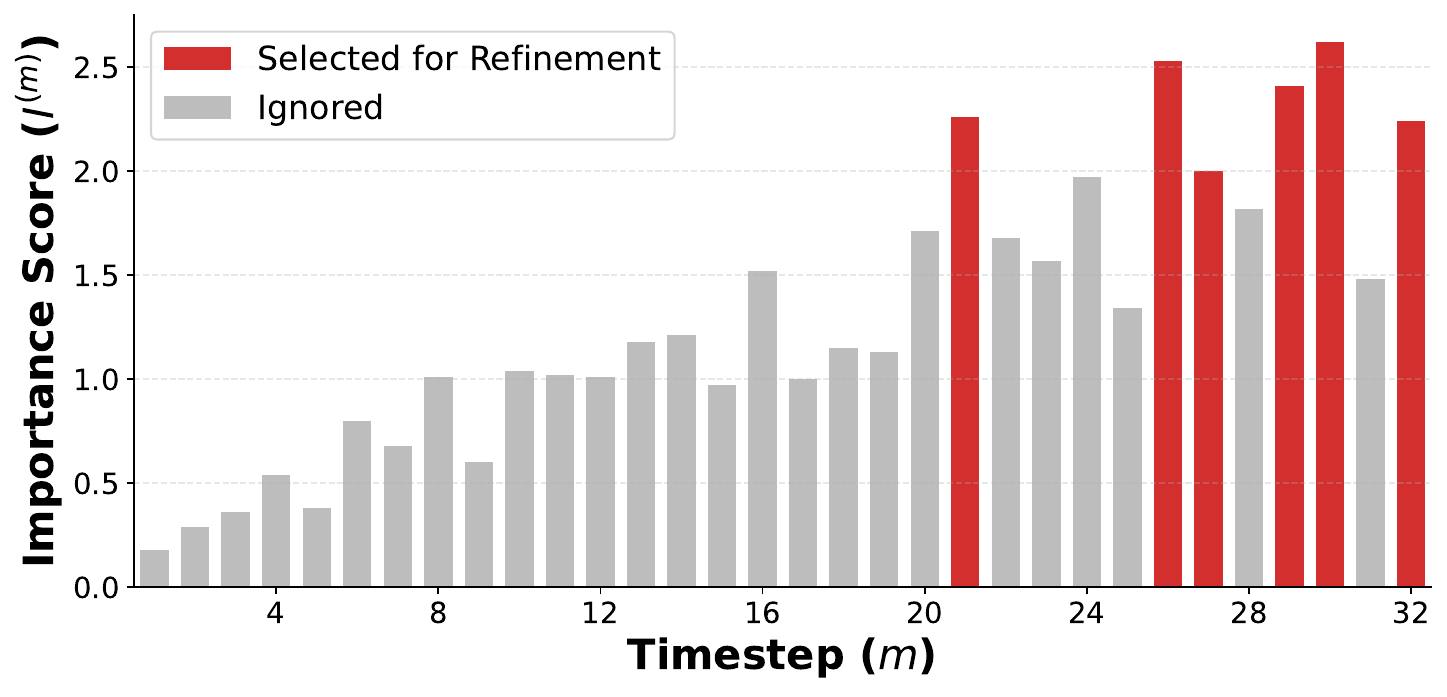}
\caption{SHAP-guided timestep selection. Red bars are selected for refinement; gray bars are ignored.}
\label{fig:shap_timestep_selection}
\vspace{-1.1em}
\end{figure}

\paragraph{SHAP-guided neuron selection}
To avoid exhaustive branching, we rank individual neurons by their gradient-based SHAP attributions on the hidden vector. Let $\Phi^{(t)}\!\in\!\mathbb{R}^s$ denote the neuron-wise attributions for the ground-truth logit $F_j$ with respect to $\mathbf{a}^{(t)}$ at timestep $t$, where $s$ is the hidden size. We rank each cross-zero neuron $(t,r)$ by its attribution magnitude $|\Phi^{(t)}_r|$ and retain the top $\rho$ fraction of candidates, forming $\mathcal{N}_{sel}$ (Algorithm~\ref{alg:shap_selection}). As illustrated in Figure~\ref{fig:shap_timestep_selection}, the corresponding timestep-level importance highlights a small subset of recurrent layers to refine; candidates are then processed in increasing $t$ so that early refinements tighten later bounds through recurrence.

\paragraph{Propagation and recursion}
Let $\mathcal{H}_{ref}$ denote the cumulative set of split constraints already imposed along the current branch. After splitting at $(t_c, r^*)$, we re-run ABP for $k=t_c\!+\!1,\ldots,m$ under each branch constraint and then apply BBP to obtain branch-wise bounds \cite{ko2019popqorn}. We recursively process the remaining candidates in $\mathcal{N}_{sel}$ that still exhibit cross-zero bounds under the current refined state up to a depth budget $d_{\max}$ (Algorithm~\ref{alg:recursive_split}, which calls Algorithm~\ref{alg:zerosplit}).

\smallskip
\textbf{Example.} 
Continuing from the previous example, suppose SHAP ranks neuron $(1,2)$ ahead of $(2,4)$ because $|\Phi_2^{(1)}|>|\Phi_4^{(2)}|$. We therefore split neuron $(1,2)$, producing a positive branch with $l_{\text{pos},2}^{(1)}\!\leftarrow\!0$ and a negative branch with $u_{\text{neg},2}^{(1)}\!\leftarrow\!0$. After re-running ABP and BBP under the two refined branches, assume the positive branch yields $\gamma^L_j=0.23$ and $\max_{i\neq j}\gamma^U_i=0.19$, while the negative branch yields $\gamma^L_j=0.27$ and $\max_{i\neq j}\gamma^U_i=0.17$. Both branches now satisfy Eq.~\eqref{eq:verify_margin}, so the original sample is certified without ever splitting neuron $(2,4)$. This illustrates the benefit of neuron-based abstraction refinement: remove the dominant cross-zero error first, then let recurrent propagation tighten the remaining bounds.

\paragraph{Optimal Split Point for Smooth Activations}
For \relu, splitting at zero is doubly justified: zero is both the non-differentiable point and the location that makes each branch exact. Smooth activations behave differently. The cross-zero condition $l^{(t)}_r<0<u^{(t)}_r$ still flags the neuron, but for \tanh{} and \textsc{Sigmoid} the relaxation gap is governed by the change of curvature across the interval rather than by a single non-smooth point, so the tightest split is not fixed at zero and depends on how $[l,u]$ is positioned. We therefore adopt a more expensive but tighter rule that selects the split point $p$ minimizing the post-split over-approximation area. Here $\sigma$ denotes the neuron's activation function. Writing $\overline{h}_1,\underline{h}_1$ for the recomputed upper and lower envelopes on $[l, p]$ and $\overline{h}_2,\underline{h}_2$ for those on$[p,u]$, the tightness measure is
\begin{equation}
\label{eq:tightness_integral}
P(\sigma, l, u, p)
  = \int_{l}^{p} \bigl[\overline{h}_1(v) - \underline{h}_1(v)\bigr]\,dv
  + \int_{p}^{u} \bigl[\overline{h}_2(v) - \underline{h}_2(v)\bigr]\,dv,
\end{equation}
and we pick $p^{*}=\arg\min_{p\in[l,u]} P(\sigma,l,u,p)$. For \relu, this degenerates to $p^{*}=0$, recovering the zero-split; for smooth activations $p^{*}$ adapts to the interval and yields a tighter pair of envelopes than zero in general.

This rule never loosens the abstraction. The endpoints $p=l$ and $p=u$ both correspond to leaving the interval unsplit, since one sub-interval collapses to zero width and contributes no area; they therefore lie in the feasible set over which $P$ is minimized. Because $p^{*}$ attains the minimum over a domain that already contains the no-split case, the refined abstraction is at least as tight as the unrefined one, and the search falls back to the endpoints exactly when refinement yields no gain. Because gated architectures such as LSTM rely heavily on \tanh{} and \textsc{Sigmoid} gates, we apply this $p^{*}$-split rather than the zero-split in our LSTM evaluation (Section~\ref{subsec:rq4}).

\begin{algorithm}[t]
\caption{The overall refinement workflow.}
\label{alg:overall_refinement}
\begin{algorithmic}[1]
\Statex \textbf{Input:} clean input sequence $\mathbf{X}$, perturbation radius $\varepsilon$, ground-truth label $j$, selection ratio $\rho$, maximal depth $d_{\max}$
\Statex \textbf{Output:} True iff $\mathbf{X}$ is certified robust under $(\varepsilon,p)$ 
\vspace{.4em}
\State $[\mathbf{l},\mathbf{u}]\gets$ \textsc{ABP}$(\mathbf{X},\varepsilon,p)$;\;
      $[\gamma^L,\gamma^U]\gets$ \textsc{BBP}$([\mathbf{l},\mathbf{u}])$
\IfReturn{$\gamma^L_j>\max_{i\neq j}\gamma^U_i$}{True}
\State $\mathcal{N}_{sel}\gets$ \textsc{Select}$(\mathbf{X},\varepsilon,j,\rho)$
\State \Return \textsc{Divide}$(\mathbf{X},\varepsilon,j,0,d_{\max},\mathcal{N}_{sel},\text{None})$
\end{algorithmic}
\end{algorithm}

\begin{algorithm}[t]
\caption{\textsc{Divide}: depth-limited recursive splitting.}
\label{alg:recursive_split}
\begin{algorithmic}[1]
\Statex \textbf{Input:} clean input sequence $\mathbf{X}$, perturbation radius $\varepsilon$, ground-truth label $j$, current depth $d$, maximal depth $d_{\max}$, selected neurons $\mathcal{N}_{sel}$, refined state $\mathcal{H}_{ref}$ (i.e., the current set of split constraints)
\Statex \textbf{Output:} True iff all induced branches can be certified (within depth budget) 
\vspace{.4em}
\IfReturn{$d\ge d_{\max}$ \textbf{or} $\mathcal{N}_{sel}$ is empty}{False}
\State $(t_c,r^*)\gets$ first $(t,r)\in\mathcal{N}_{sel}$ such that $l^{(t)}_{r}<0<u^{(t)}_{r}$ under $\mathcal{H}_{ref}$
\State $(\gamma_{pos},\gamma_{neg},\mathcal{H}_{pos},\mathcal{H}_{neg})\gets$ \textsc{Split}$(\mathbf{X},\varepsilon,p,(t_c,r^*),\mathcal{H}_{ref})$
\IfReturn{\textsc{Verify}($\gamma_{pos}$) \textbf{and} \textsc{Verify}($\gamma_{neg}$)}{True}
\State \Return \textsc{Divide}$(...,d\!+\!1,\mathcal{H}_{pos})\wedge$\textsc{Divide}$(...,d\!+\!1,\mathcal{H}_{neg})$
\end{algorithmic}
\end{algorithm}

\begin{algorithm}[t]
\caption{\textsc{Select}: critical neuron selection.}
\label{alg:shap_selection}
\begin{algorithmic}[1]
\Statex \textbf{Input:} clean input sequence $\mathbf{X}$, perturbation radius $\varepsilon$, ground-truth label $j$, selection ratio $\rho$
\Statex \textbf{Output:} $\mathcal{N}_{sel}$: neurons $\{(t,r)\}$ sorted ascending by $t$
\vspace{.4em}
\State Sample background $\mathcal{B}\subset\mathbb{B}_p(\mathbf{X},\varepsilon)$
\State $m\gets$ sequence length of $\mathbf{X}$
\For{$t=1$ to $m$}
  \State Compute $\Phi^{(t)}\gets$ Grad-SHAP on hidden vector $\mathbf{a}^{(t)}$
  \For{each neuron $r$}
    \If{$l^{(t)}_r < 0 < u^{(t)}_r$}
      \State Add $(t,\,r,\,|\Phi^{(t)}_r|)$ to candidate list $\mathcal{C}$
    \EndIf
  \EndFor
\EndFor
\IfReturn{$\mathcal{C}=\varnothing$}{$\varnothing$}
\State $K\gets \max(1,\lceil \rho\cdot |\mathcal{C}|\rceil)$
\State \Return top-$K$ neurons $(t,r)$ from $\mathcal{C}$ ranked by $|\Phi^{(t)}_r|$, sorted ascending by $t$
\end{algorithmic}
\end{algorithm}

\begin{algorithm}[t]
\caption{\textsc{Split}: branch generation and propagation.}
\label{alg:zerosplit}
\begin{algorithmic}[1]
\Statex \textbf{Input:} clean input sequence $\mathbf{X}$, perturbation radius $\varepsilon$, critical neuron $(t_c,r^*)$, refined state $\mathcal{H}_{ref}$ (i.e., the current set of split constraints)
\Statex \textbf{Output:} branch bounds $\gamma_{pos},\gamma_{neg}$ and updated refined states $\mathcal{H}_{pos},\mathcal{H}_{neg}$
\vspace{.4em}
\State Apply $\mathcal{H}_{ref}$ to current $[\mathbf{l},\mathbf{u}]$ and copy bounds to pos/neg
\State Set $l^{(t_c)}_{\text{pos},r^*}\!\leftarrow\!0$ and $u^{(t_c)}_{\text{neg},r^*}\!\leftarrow\!0$
\State Re-run ABP for $k=t_c\!+\!1,\ldots,m$ in each branch
\State Run BBP to get $\gamma_{pos},\gamma_{neg}$
\State \Return $(\gamma_{pos},\gamma_{neg},\mathcal{H}_{pos},\mathcal{H}_{neg})$
\end{algorithmic}

\end{algorithm}

\section{Experiment Evaluation}
\label{c:experiment}

We evaluate our method on two RNN and one LSTM settings, addressing four research questions:
(RQ1) certification gains over abstraction-only verification,
(RQ2) runtime overhead of abstraction refinement across activation types and sequence lengths,
(RQ3) computational cost of SHAP-guided neuron selection, and
(RQ4) effectiveness of abstraction refinement when extended to LSTMs.

\subsection{Experiment Setup}
\label{sec:setup}

\textit{Datasets, models, and property.}
We train RNNs of hidden layer sizes $h\in\{16,32,64,128\}$ on CIFAR10, MNIST Stroke sequences~\cite{liwicki2012recognition}, and MNIST. CIFAR10 images are reshaped into sequences with $m\in\{8,12,24,32\}$, MNIST Strokes have sequence lengths $m\in\{30,35,40,45\}$, and the LSTM extension uses MNIST models with hidden size $h\in\{4,8,16,32\}$ and $m\in\{1,2,4,7\}$. For each clean input $\mathbf{X}_0$ with $j=\arg\max_i F_i(\mathbf{X}_0)$, we certify top-1 label preservation under per-timestep $L_2$ perturbations, i.e., $F_j(\mathbf{X})>F_i(\mathbf{X})$ for all $i\neq j$ and all $\mathbf{X}$ in the $\varepsilon$-ball at each step. We use the first 50 samples from each dataset and apply refinement without filtering based on prediction correctness.

\textit{Metrics and refinement policy.}
Following prior work~\cite{li2023towards}, we use an EVR-style sweep in which $\varepsilon$ ranges from 0.005 to 0.1 in increments of 0.001. The abstraction-only verifier runs until it first fails; refinement is then invoked, and a sample is counted as recovered ($\uparrow$) if the refined abstraction certifies it at that $\varepsilon$. In Tables~\ref{tab:evr_cifar10}, ~\ref{tab:evr_seq} and~\ref{tab:evr_lstm}, $\times$ denotes the number of samples that trigger refinement during the sweep, and $\uparrow$ denotes the recovered subset. Candidate neurons are ranked by a gradient-based explainer~\cite{sundararajan2017axiomaticattributiondeepnetworks} on hidden states with randomly perturbed backgrounds, filtered to cross-zero neurons, and truncated to the top five candidates. Each recursive split uses the first temporally valid candidate that has not been split before and still crosses zero under the refined bounds, with at most 31 refinements per configuration. All models, datasets, scripts, and reproduction instructions are available at \url{https://github.com/leolin-hub/ZeroSplitVerifier}.

\subsection{Certification Success Rate and Model Sensitivity (RQ1)}

Tables~\ref{tab:evr_cifar10} and~\ref{tab:evr_seq} show that abstraction refinement improves certification in almost all settings, recovering additional robust samples that abstraction alone misses.
The dominant factor is sequence length: on both CIFAR10 and MNIST Stroke, aggregate recovery decreases as $m$ grows because a fixed split budget removes a smaller fraction of the relaxation error accumulated across a longer recurrence. Width introduces a related but less monotone effect. Wider models create more cross-zero candidates, so the verifier must identify a smaller set of influential neurons from a larger search space; this makes recovery harder in aggregate, although {\relu} sometimes peaks at medium hidden size when the added capacity improves the baseline margins without overwhelming the split budget. The activation trend follows the same mechanism. In narrow models, {\tanh} can benefit substantially because tightening a few influential smooth envelopes can move the final margin, whereas in wider models {\relu} becomes more favorable because splitting makes the selected neuron exact rather than merely tighter.

\begin{table}[t]
\centering
\caption{EVR improvements on CIFAR10}
\label{tab:evr_cifar10}
\renewcommand{\arraystretch}{0.80}
\scriptsize
\begin{tabular*}{\columnwidth}{@{\extracolsep{\fill}}ccccccccccc@{}}
\toprule
$m$ & Act. & \multicolumn{2}{c}{$h=16$} & \multicolumn{2}{c}{$h=32$} & \multicolumn{2}{c}{$h=64$} & \multicolumn{2}{c}{$h=128$} \\
\cmidrule{3-10}
& & $\times$ & $\uparrow$ & $\times$ & $\uparrow$ & $\times$ & $\uparrow$ & $\times$ & $\uparrow$ \\
\midrule
\multirow{2}{*}{8}  & \relu  & 43 & 24 & 47 & 33 & 50 & 32 & 50 & 25 \\
                    & \tanh  & 50 & 35 & 50 & 28 & 50 & 30 & 50 & 25 \\
\midrule
\multirow{2}{*}{12} & \relu  & 47 & 16 & 50 & 20 & 50 & 26 & 50 & 22 \\
                    & \tanh  & 50 & 34 & 50 & 27 & 50 & 15 & 50 & 11 \\
\midrule
\multirow{2}{*}{24} & \relu  & 50 & 16 & 50 & 19 & 50 & 12 & 50 & 8 \\
                    & \tanh  & 50 & 21 & 50 & 13 & 50 & 11 & 50 & 1 \\
\midrule
\multirow{2}{*}{32} & \relu  & 50 & 18 & 50 & 17 & 50 & 12 & 50 & 4 \\
                    & \tanh  & 50 & 16 & 50 & 10 & 50 & 4 & 50 & 1 \\
\bottomrule
\end{tabular*}
\end{table}
\begin{table}[t]
\centering
\caption{EVR improvements on MNIST Strokes}
\label{tab:evr_seq}
\renewcommand{\arraystretch}{0.80}
\scriptsize
\begin{tabular*}{\columnwidth}{@{\extracolsep{\fill}}ccccccccccc@{}}
\toprule
$m$ & Act. & \multicolumn{2}{c}{$h=16$} & \multicolumn{2}{c}{$h=32$} & \multicolumn{2}{c}{$h=64$} & \multicolumn{2}{c}{$h=128$} \\
\cmidrule{3-10}
& & $\times$ & $\uparrow$ & $\times$ & $\uparrow$ & $\times$ & $\uparrow$ & $\times$ & $\uparrow$ \\
\midrule
\multirow{2}{*}{30}  & \relu  & 50 & 20 & 50 & 24 & 50 & 10 & 50 & 4 \\
                    & \tanh  & 50 & 30 & 50 & 19 & 50 & 9 & 50 & 3 \\
\midrule
\multirow{2}{*}{35} & \relu  & 50 & 15 & 50 & 17 & 50 & 6 & 50 & 4 \\
                    & \tanh  & 50 & 21 & 50 & 11 & 50 & 4 & 50 & 0 \\
\midrule
\multirow{2}{*}{40} & \relu  & 50 & 13 & 50 & 17 & 50 & 7 & 50 & 3 \\
                    & \tanh  & 50 & 22 & 50 & 2 & 50 & 2 & 50 & 0 \\
\midrule
\multirow{2}{*}{45} & \relu  & 50 & 5 & 50 & 11 & 50 & 3 & 50 & 6 \\
                    & \tanh  & 50 & 16 & 50 & 9 & 50 & 6 & 50 & 1 \\
\bottomrule
\end{tabular*}
\end{table}

\subsection{Runtime Overhead vs Abstraction Baseline (RQ2)}

\begin{figure}[t]
    \centering
    \includegraphics[width=\linewidth]{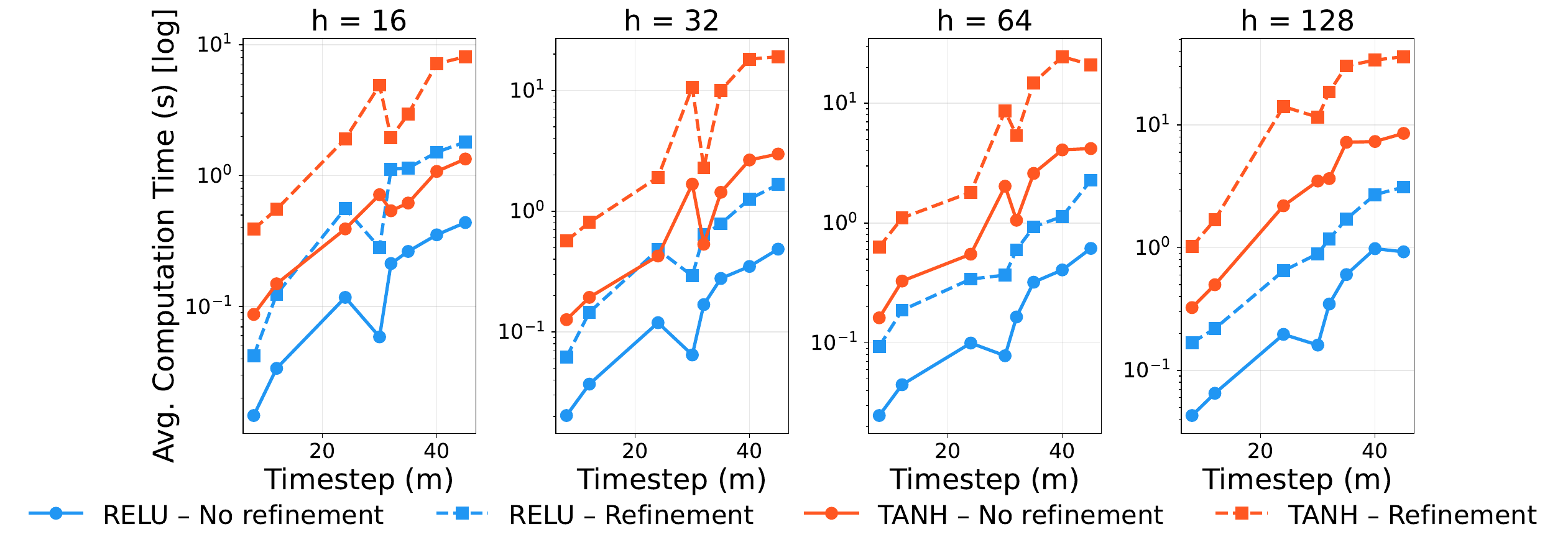}
    \caption{Average computation time (log scale) of refinement and the baseline across $m$ under different $h$; the baseline is the abstraction-only runtime.}
    \label{fig:m_scaling}
\end{figure}

\begin{figure}[t]
    \centering
    \includegraphics[width=0.8\linewidth]{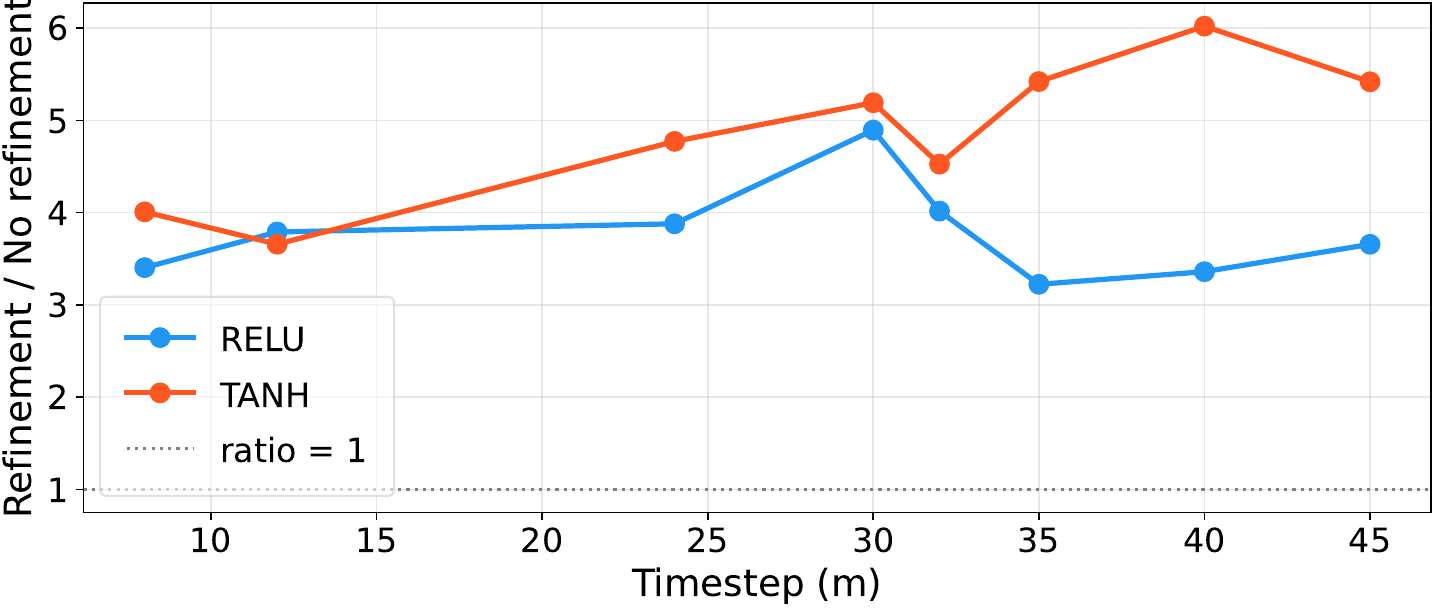}
    \caption{Computation time ratio of refinement vs.~no refinement. Note that no refinement denotes the abstraction-only approach.}
    \label{fig:overhead_ratio}
\end{figure}

Figures~\ref{fig:m_scaling} and~\ref{fig:overhead_ratio} compare refinement with the abstraction-only baseline. Figure~\ref{fig:m_scaling} reports average verification time over the full $\varepsilon$ sweep, while Figure~\ref{fig:overhead_ratio} normalizes this cost by the abstraction-only runtime after averaging over hidden sizes. Both methods become slower as $m$ increases because ABP/BBP must traverse more recurrent steps; refinement amplifies this scaling because each selected split creates branches whose downstream bounds must be recomputed. The activation-dependent overhead is therefore not just an implementation artifact: {\tanh} remains consistently more expensive because smooth envelopes must be recomputed in each branch, whereas a split {\relu} neuron becomes exact. The ratio varies with $m$ as this branch cost is amortized differently across configurations, but it remains above $1$, confirming that refinement trades additional computation for tighter certificates rather than providing an end-to-end speedup.

\subsection{Efficiency vs. Refinement Effectiveness (RQ3)}
\begin{figure}[t]
    \centering
    \includegraphics[width=0.82\linewidth]{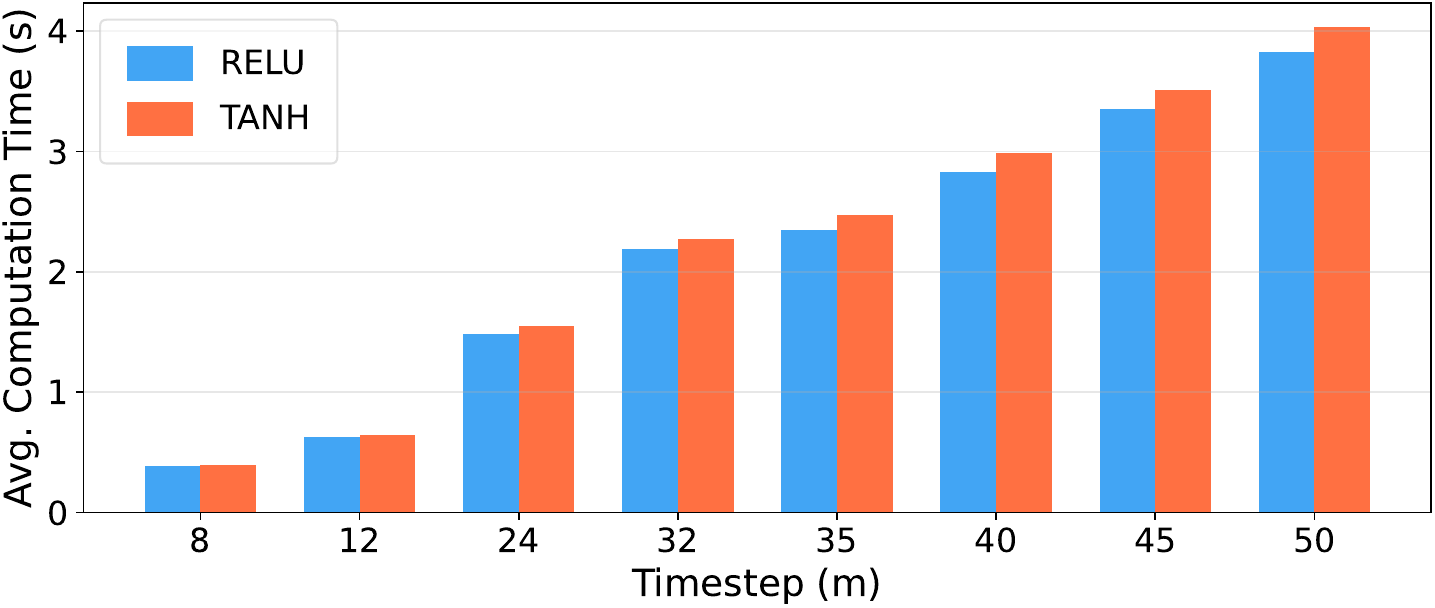}
    \caption{Computation time for SHAP scores.}
    \label{fig:shap_time}
\end{figure}

Figure~\ref{fig:shap_time} shows that SHAP scoring grows steadily with sequence length, while the {\relu} and {\tanh} curves remain close. This matches the role of SHAP in our pipeline: attribution requires backpropagating through the unrolled recurrent computation, so its cost is governed primarily by sequence length, whereas the activation-specific expense appears later during branch verification and envelope recomputation. Because SHAP is paid once per sample, it acts as a front-end filtering cost that is amortized over subsequent refinement calls. Combined with RQ1 and RQ2, this indicates that the main computational burden is not candidate scoring itself but verifying the selected branches; SHAP is useful precisely because it directs that more expensive verification budget toward the neurons most likely to affect the final margin.

\subsection{Effectiveness of Abstraction Refinement on LSTMs (RQ4)}
\label{subsec:rq4}
\begin{table}[t]
\centering
\caption{EVR improvements on LSTM over MNIST}
\label{tab:evr_lstm}
\renewcommand{\arraystretch}{0.8}
\scriptsize
\begin{tabular*}{\columnwidth}{@{\extracolsep{\fill}}ccccccccccc@{}}
\toprule
$m$ & \multicolumn{2}{c}{$h=4$} & \multicolumn{2}{c}{$h=8$}
    & \multicolumn{2}{c}{$h=16$} & \multicolumn{2}{c}{$h=32$} \\
\cmidrule(lr){2-3}\cmidrule(lr){4-5}\cmidrule(lr){6-7}\cmidrule(lr){8-9}
& $\times$ & $\uparrow$ & $\times$ & $\uparrow$
& $\times$ & $\uparrow$ & $\times$ & $\uparrow$ \\
\midrule
1 & 44 & 11 & 40 & 22 & 27 & 23 & 16 & 14 \\
2 & 50 & 19 & 49 & 38 & 43 & 32 & 34 & 23 \\
4 & 50 & 24 & 50 & 23 & 50 & 24 & 50 & 18 \\
7 & 50 & 11 & 50 & 15 & 50 & 8  & 50 & 8 \\
\bottomrule
\end{tabular*}
\end{table}

We evaluate the refinement strategy on small LSTM models over MNIST to test whether it extends to gated architectures, applying the $p^{*}$-split introduced in Section~\ref{c:zerosplit} at the \textsc{Sigmoid} and \tanh\ gates. Table~\ref{tab:evr_lstm} shows that refinement recovers additional certified samples in every configuration. As the sequence length $m$ grows, almost all samples fail the baseline and trigger refinement. The number of recovered samples follows a rise-then-fall pattern, peaking at $m=2$ and dropping to its lowest at $m=7$. The rise reflects a growing candidate pool: as $m$ increases more samples fail the baseline and enter refinement, and at $m=2$ a single split can still tighten most of the accumulated relaxation, giving the greatest recovery. Beyond that the pool saturates but each added timestep stacks four more gate relaxations, so one split removes a steadily smaller share of the total error and recovery declines.
Across model width, the number of recovered samples peaks at intermediate hidden sizes around $h=8$ and tapers for the widest model. At $h=4$ the candidate pool is too small for SHAP-guided selection to locate a neuron whose split meaningfully tightens the bound, whereas at larger widths each split removes a smaller share of the total relaxation error and the baseline abstraction is already tighter. Overall, targeted refinement carries over to LSTMs, recovering the most samples at short-to-intermediate sequence lengths and intermediate widths, with recovery tapering off for long sequences and the widest model owing to the per-step gate nonlinearities.

\subsection{Threats to Validity}
\label{sec:threats}

\paragraph{Experimental scope and protocol}
Our evaluation covers two main sequence benchmarks and a small LSTM extension, exercising both image-as-sequence and trajectory-style inputs but not all recurrent architectures, longer horizons, larger hidden states, or tasks with stronger temporal dependencies. We also use the first 50 test samples from each dataset for reproducibility and tractability; this subset may not fully represent the complete test distribution, and larger stratified or randomly sampled test sets with confidence intervals would better quantify recovery rates. The reported values are tied to certified top-1 preservation under per-timestep $L_2$ perturbations and a fixed $\varepsilon$ sweep; other perturbation models, temporally correlated noise, $\varepsilon$ schedules, or protocols such as refining at every radius or enforcing a global time budget could change the observed runtime-recovery trade-off.

\paragraph{Sensitivity and randomness}
Refinement depends on gradient-based SHAP scores computed from randomly perturbed background samples, the candidate selection ratio, the depth budget, and the rule for choosing the next temporally valid cross-zero neuron. Different background sample sizes, background draws, or random seeds can change attribution rankings, especially when several neurons have similar scores. Larger candidate/depth budgets may recover more certificates at higher branching cost, while smaller budgets may miss useful split locations. Training randomness, initialization, and data ordering can also change the learned recurrent dynamics, abstraction tightness, cross-zero intervals, and SHAP rankings; repeating experiments over multiple trained checkpoints would better estimate model-to-model variability.

\paragraph{Algorithmic limitations}
The procedure is sound but incomplete: failure after refinement means only that the current abstraction and budget did not prove robustness, not that the input is non-robust. Worst-case branching remains exponential, and practical effectiveness depends on whether the selected neurons remove enough relaxation error to change the final margin. Runtime is implementation-dependent as well; batching, caching, or parallel branch evaluation could alter absolute overhead while preserving the qualitative trade-off between tighter certificates and additional computation.

\section{Related Work and Concluding Remarks}
\label{c:related}
\label{sec:conclusion}

\paragraph{Robustness verification}
Adversarial attacks~\cite{szegedy2013intriguing, madry2017towards} motivate certified verification. For feed-forward networks, scalable abstractions use intervals~\cite{gehr2018ai2, zhang2018efficient}, zonotopes~\cite{singh2018fast}, and polyhedra~\cite{singh2019abstract}, while SMT/MILP solvers such as \textsc{Reluplex} and \textsc{Marabou} provide more complete but less scalable reasoning~\cite{katz2017reluplex, katz2019marabou}. RNN verification inherits these ideas but must handle recurrent unrolling, where relaxation error at one timestep can propagate and compound. \textsc{Popqorn}~\cite{ko2019popqorn} extends linear bound propagation to RNNs and serves as our abstraction baseline; other RNN certifiers use interval-style abstractions~\cite{du2021cert, guo2021rnnguard, paulsen2020arc}, differential verification~\cite{mohammadinejad2021diffrnn}, or reachability/star-set analysis~\cite{bak2020venus}. These methods improve the abstraction itself but do not target which recurrent states should be refined when cross-zero relaxations dominate the certification gap.

\paragraph{Refinement and branching}
Refinement tightens failed abstractions by conditioning on additional constraints. For FNNs, branch-and-bound splits activation domains~\cite{bunel2020branch, wang2021beta, depalma2021improved}, cutting-plane methods strengthen relaxations~\cite{zhou2025scalable}, and debugging-oriented refinement identifies high-impact zero-crossing regions~\cite{singh2018fast}. Sequential refinement has also appeared through gradient-guided polyhedral RNN verification~\cite{ryou2021scalable}, branching for general nonlinearities~\cite{shi2024neural}, and adaptive reachability refinement~\cite{sun2022marble}. These approaches either refine broadly or optimize local branches without prioritizing timesteps under recurrent error propagation. Our method instead uses SHAP-style attribution~\cite{lundberg2017unified}, implemented with gradient-based computation~\cite{sundararajan2017axiomaticattributiondeepnetworks}, to select the hidden neurons and timesteps most likely to improve the final robustness margin.

\paragraph{Concluding remarks}
This work introduces a SHAP-guided abstraction-refinement framework for certified local robustness verification of RNNs: refining influential cross-zero hidden neurons tightens output bounds and recovers certificates missed by abstraction-only propagation. Results show consistent gains on vanilla RNNs and an extension to LSTMs. Future work includes predictive split policies that estimate margin gain before branching, verifier-aware training to reduce high-impact cross-zero states, and extensions to GRUs, encoder--decoder models, streaming classifiers, and event-level robustness properties.



\begin{thebibliography}{10}
\providecommand{\url}[1]{#1}
\csname url@samestyle\endcsname
\providecommand{\newblock}{\relax}
\providecommand{\bibinfo}[2]{#2}
\providecommand{\BIBentrySTDinterwordspacing}{\spaceskip=0pt\relax}
\providecommand{\BIBentryALTinterwordstretchfactor}{4}
\providecommand{\BIBentryALTinterwordspacing}{\spaceskip=\fontdimen2\font plus
\BIBentryALTinterwordstretchfactor\fontdimen3\font minus
  \fontdimen4\font\relax}
\providecommand{\BIBforeignlanguage}[2]{{%
\expandafter\ifx\csname l@#1\endcsname\relax
\typeout{** WARNING: IEEEtran.bst: No hyphenation pattern has been}%
\typeout{** loaded for the language `#1'. Using the pattern for}%
\typeout{** the default language instead.}%
\else
\language=\csname l@#1\endcsname
\fi
#2}}
\providecommand{\BIBdecl}{\relax}
\BIBdecl

\bibitem{sutskever2014sequence}
I.~Sutskever, O.~Vinyals, and Q.~V. Le, ``Sequence to sequence learning with
  neural networks,'' in \emph{Proc. Adv. Neural Inf. Process. Syst. (NeurIPS)},
  2014, pp. 3104--3112.

\bibitem{graves2013speech}
A.~Graves, A.-r. Mohamed, and G.~Hinton, ``Speech recognition with deep
  recurrent neural networks,'' in \emph{Proc. IEEE Int. Conf. Acoust., Speech
  Signal Process. (ICASSP)}, 2013, pp. 6645--6649.

\bibitem{nelson2017stock}
D.~M. Nelson, A.~C. Pereira, and R.~A. de~Oliveira, ``Stock market's price
  movement prediction with lstm neural networks,'' in \emph{Proc. Int. Joint
  Conf. Neural Netw. (IJCNN)}, 2017, pp. 1419--1426.

\bibitem{alahi2016social}
A.~Alahi, K.~Goel, V.~Ramanathan, A.~Robicquet, L.~Fei-Fei, and S.~Savarese,
  ``Social lstm: Human trajectory prediction in crowded spaces,'' in
  \emph{Proc. IEEE Conf. Comput. Vis. Pattern Recognit. (CVPR)}, 2016, pp.
  961--971.

\bibitem{szegedy2013intriguing}
C.~Szegedy, W.~Zaremba, I.~Sutskever, J.~Bruna, D.~Erhan, I.~Goodfellow, and
  R.~Fergus, ``Intriguing properties of neural networks,''
  \emph{arXiv:1312.6199}, 2013.

\bibitem{goodfellow2014explaining}
I.~J. Goodfellow, J.~Shlens, and C.~Szegedy, ``Explaining and harnessing
  adversarial examples,'' in \emph{Proc. Int. Conf. Learn. Represent. (ICLR)},
  2015.

\bibitem{jia2021fooling}
Y.~Jia, J.~Lu, J.~Goswami, H.~Li, X.~Li, and B.~Zhao, ``Fooling lidar
  perception via adversarial trajectory perturbation,'' in \emph{Proc. IEEE/CVF
  Int. Conf. Comput. Vis. (ICCV)}, 2021, pp. 7898--7907.

\bibitem{ko2019popqorn}
C.-Y. Ko, D.~Gopinath, M.~Shih, C.~Pasareanu, and S.~Khurshid, ``Popqorn:
  Quantifying robustness of recurrent neural networks,'' in \emph{Proc. Int.
  Conf. Mach. Learn. (ICML)}, 2019, pp. 3468--3477.

\bibitem{li2023towards}
J.~Li, G.~Bai, L.~H. Pham, and J.~Sun, ``Towards an effective and interpretable
  refinement approach for dnn verification,'' in \emph{Proc. IEEE Int. Conf.
  Softw. Qual., Rel., Security (QRS)}.\hskip 1em plus 0.5em minus 0.4em\relax
  IEEE, 2023, pp. 569--580.

\bibitem{liwicki2012recognition}
M.~Liwicki, A.~Graves, H.~Bunke, and J.~Schmidhuber, ``Recognition of
  whiteboard notes: Online, offline and combination,'' in \emph{Ser. Mach.
  Percept. Artif. Intell.}, vol.~75.\hskip 1em plus 0.5em minus 0.4em\relax
  World Scientific, 2012, pp. 1--13.

\bibitem{sundararajan2017axiomaticattributiondeepnetworks}
\BIBentryALTinterwordspacing
M.~Sundararajan, A.~Taly, and Q.~Yan, ``Axiomatic attribution for deep
  networks,'' 2017. [Online]. Available: \url{https://arxiv.org/abs/1703.01365}
\BIBentrySTDinterwordspacing

\bibitem{madry2017towards}
A.~Madry, A.~Makelov, L.~Schmidt, D.~Tsipras, and A.~Vladu, ``Towards deep
  learning models resistant to adversarial attacks,'' \emph{arXiv:1706.06083},
  2017.

\bibitem{gehr2018ai2}
T.~Gehr, M.~Mirman, D.~Drachsler-Cohen, P.~Tsankov, S.~Chaudhuri, and
  M.~Vechev, ``Ai2: Safety and robustness certification of neural networks with
  abstract interpretation,'' in \emph{Proc. IEEE Symp. Secur. Privacy (SP)},
  2018, pp. 3--18.

\bibitem{zhang2018efficient}
H.~Z. Zhang, T.-W. Weng, P.-Y. Chen, C.-J. Xu, and C.-J. Hsieh, ``Efficient
  neural network robustness certification with general activation functions,''
  in \emph{Proc. Adv. Neural Inf. Process. Syst. (NeurIPS)}, 2018, pp.
  4939--4948.

\bibitem{singh2018fast}
G.~Singh, T.~Gehr, M.~P{\"u}schel, and M.~Vechev, ``Fast and precise
  certification of neural networks,'' in \emph{Proc. Adv. Neural Inf. Process.
  Syst. (NeurIPS)}, 2018, pp. 10\,802--10\,813.

\bibitem{singh2019abstract}
G.~Singh, T.~Gehr, M.~Mirman, M.~P{\"u}schel, and M.~Vechev, ``An abstract
  domain for certifying neural networks,'' in \emph{Proc. ACM Program. Lang.
  (POPL)}, vol.~3, no. POPL, 2019, pp. 1--30.

\bibitem{katz2017reluplex}
G.~Katz, C.~Barrett, D.~L. Dill, K.~Julian, and M.~J. Kochenderfer, ``Reluplex:
  An efficient smt solver for verifying deep neural networks,'' in \emph{Proc.
  Int. Conf. Comput. Aided Verification (CAV)}.\hskip 1em plus 0.5em minus
  0.4em\relax Springer, 2017, pp. 97--117.

\bibitem{katz2019marabou}
------, ``The marabou framework for verification and analysis of deep neural
  networks,'' in \emph{Proc. Int. Conf. Comput. Aided Verification
  (CAV)}.\hskip 1em plus 0.5em minus 0.4em\relax Springer, 2019, pp. 443--452.

\bibitem{du2021cert}
M.~Du, H.~Zhang, S.~Wang, H.~Jin, and C.-J. Hsieh, ``Cert-rnn: Towards
  certifying the robustness of recurrent neural networks,'' in \emph{Proc. IEEE
  Symp. Secur. Privacy (SP)}, 2021, pp. 968--985.

\bibitem{guo2021rnnguard}
T.~Guo, K.~Xu, H.~Zhang, and D.~Lin, ``Rnn-guard: Certified robustness against
  adversarial examples in recurrent neural networks,'' in \emph{Proc. Int.
  Conf. Learn. Represent. (ICLR)}, 2021.

\bibitem{paulsen2020arc}
B.~Paulsen, J.~Wang, and C.~Wang, ``Certified robustness to programmable
  transformations in lstms,'' in \emph{Proc. Eur. Symp. Res. Comput. Security
  (ESORICS)}.\hskip 1em plus 0.5em minus 0.4em\relax Springer, 2020, pp.
  562--582.

\bibitem{mohammadinejad2021diffrnn}
S.~Mohammadinejad, B.~Paulsen, J.~V. Deshmukh, and C.~Wang, ``Diffrnn:
  Differential verification of recurrent neural networks,'' in \emph{FORMATS},
  2021.

\bibitem{bak2020venus}
S.~Bak, C.~Liu, and T.~Johnson, ``Verification of recurrent neural networks
  using star reachability,'' in \emph{Proc. Int. Conf. Comput. Aided
  Verification (CAV)}.\hskip 1em plus 0.5em minus 0.4em\relax Springer, 2020,
  pp. 341--363.

\bibitem{bunel2020branch}
R.~Bunel, J.~M. Lu, I.~Turkaslan, P.~H. Torr, P.~Kohli, and M.~P. Kumar,
  ``Branch and bound for piecewise linear neural network verification,''
  \emph{J. Mach. Learn. Res.}, vol.~21, no.~42, pp. 1--39, 2020.

\bibitem{wang2021beta}
S.~Wang, H.~Zhang, K.~Xu, X.~Chen, D.~Lin, S.~Jana, J.~Z. Kolter, and E.~Wong,
  ``Beta-crown: Efficient bound propagation with per-neuron split constraints
  for neural network verification,'' in \emph{Proc. Adv. Neural Inf. Process.
  Syst. (NeurIPS)}, 2021.

\bibitem{depalma2021improved}
A.~De~Palma, R.~Bunel, A.~Desmaison, K.~Dvijotham, P.~Kohli, P.~H. Torr, and
  M.~P. Kumar, ``Improved branch and bound for neural network verification via
  lagrangian decomposition,'' \emph{arXiv:2104.06718}, 2021.

\bibitem{zhou2025scalable}
D.~Zhou, C.~Brix, G.~A. Hanasusanto, and H.~Zhang, ``Scalable neural network
  verification with branch-and-bound inferred cutting planes,'' in
  \emph{NeurIPS}, 2024.

\bibitem{ryou2021scalable}
W.~Ryou, J.~Chen, M.~Balunovic, G.~Singh, A.~Dan, and M.~Vechev, ``Scalable
  polyhedral verification of recurrent neural networks,'' in \emph{Proc. Adv.
  Neural Inf. Process. Syst. (NeurIPS)}, 2020.

\bibitem{shi2024neural}
Z.~Shi, Q.~Jin, Z.~Kolter, S.~Jana, C.-J. Hsieh, and H.~Zhang, ``Neural network
  verification with branch-and-bound for general nonlinearities,'' in
  \emph{TACAS}, 2025.

\bibitem{sun2022marble}
X.~Sun, H.~Kazemi, Z.~Xu, and C.~Wang, ``Marble: Model-based robustness
  analysis of stateful deep learning systems,'' in \emph{Proc. 37th IEEE/ACM
  Int. Conf. Autom. Softw. Eng. (ASE)}, 2022, pp. 1--13.

\bibitem{lundberg2017unified}
S.~M. Lundberg and S.-I. Lee, ``A unified approach to interpreting model
  predictions,'' in \emph{Proc. Adv. Neural Inf. Process. Syst.}, 2017, pp.
  4765--4774.

\end{thebibliography}
\end{document}